\documentclass[10pt,twocolumn,letterpaper]{article}

 \usepackage{cvpr}              
\usepackage{graphicx}
\usepackage{amsmath}
\usepackage{amssymb}
\usepackage{booktabs}
\usepackage{bm}
\usepackage{multirow}
\usepackage{makecell}

\usepackage[pagebackref,breaklinks,colorlinks,citecolor=cvprblue]{hyperref}

\def\paperID{13873} 
\def\confName{CVPR}
\def\confYear{2024}

\title{Diverse Image Harmonization}

\author{Xinhao Tao, Tianyuan Qiu, Junyan Cao, Li Niu\thanks{Corresponding author.} \\
MoE Key Lab of Artificial Intelligence, Shanghai Jiao Tong University\\
\{taoxinhao, frank\_qiu, joy\_c1,ustcnewly\}@sjtu.edu.cn
}

\begin{document}
\maketitle
\definecolor{cvprblue}{rgb}{0.21,0.49,0.74}
\begin{abstract}
Image harmonization aims to adjust the foreground illumination in a composite image to make it harmonious. The existing harmonization methods can only produce one deterministic result for a composite image, ignoring that a composite image could have multiple plausible harmonization results due to multiple plausible reflectances. In this work, we first propose a reflectance-guided harmonization network, which can achieve better performance with the guidance of ground-truth foreground reflectance. Then, we also design a diverse reflectance generation network to predict multiple plausible foreground reflectances, leading to multiple plausible harmonization results. The extensive experiments on the benchmark datasets demonstrate the effectiveness of our method. 
\end{abstract}

\section{Introduction}

Image composition \cite{imagecomposite2} refers to combining visual elements from different images into a realistic composite image, which has diverse applications from everyday photo editing to automatic advertising. To tackle the illumination discrepancy between foreground and background, image harmonization aims to adjust the illumination of foreground to produce a harmonious image. 
Specifically, given a composite image $\bm{I}_c$ and foreground mask $\bm{M}$, image harmonization model produces a harmonization result. Recently, deep learning based image harmonization has made huge progress. The existing methods formulate image harmonization as image translation \cite{dovenet, IIH, ssh, iSSAM} or color transformation \cite{CDTNet, DCCF, S2CRNet, PCT-Net}, producing visually pleasant results. \par

\begin{figure}[t]
\centering
\includegraphics[width=1.0\linewidth]{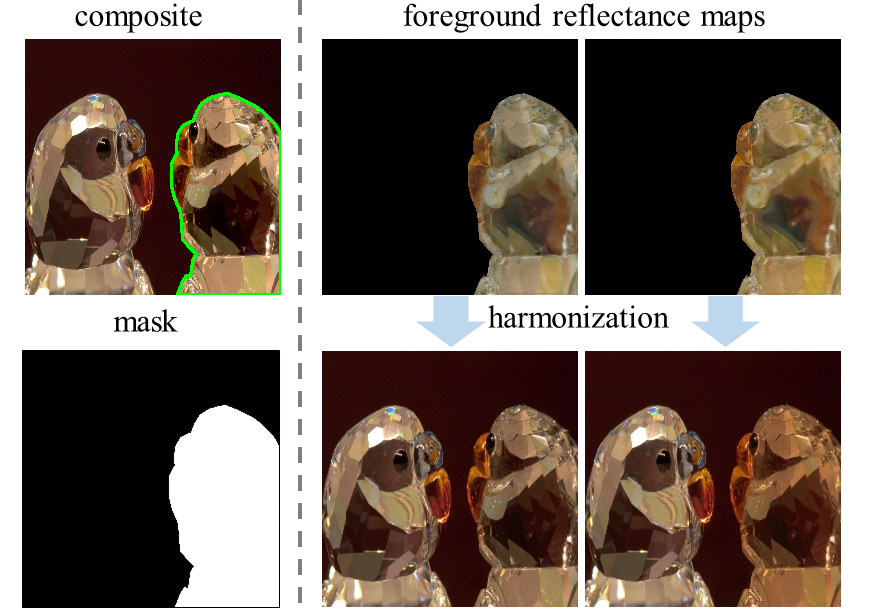}
\caption{An example of multiple plausible harmonization results when the foreground reflectance is uncertain.}
\label{fig:example}
\end{figure}

All the above methods can only produce one deterministic result for a composite image. However, a composite image may actually have multiple plausible harmonization results. According to the Retinex theory~\cite{kimmel2003variational},
an image $\bm{I}$ could be decomposed into reflectance map $\bm{A}$ and illumination map $\bm{L}$: $\bm{I} = \bm{A}*\bm{L}$, in which * means element-wise product. Intuitively, the reflectance map represents the essential color and the illumination map represents the environmental lighting. 
Given the foreground in composite image $\bm{I}_c$, it is sometimes difficult to determine its reflectance and there could be multiple plausible reflectances. Therefore, \textbf{when transferring the background illumination to the foreground, different foreground reflectances would yield different harmonization results}, which is referred to as diverse image harmonization in this paper. We provide one example of diverse image harmonization in Figure \ref{fig:example}. The composite foreground is a yellow glass figurine. It is hard to determine whether it is a white figurine under yellow illumination or a yellow figurine under white illumination. The harmonization results for these two cases should be different.

One way to disambiguate the foreground reflectance is referring to the original foreground image $\bm{I}_o$ that the foreground belongs to. \textbf{Compared with the foreground, $\bm{I}_o$ contains richer colors and semantics, which could help determine the illumination and thus further determine the foreground reflectance.} In the most commonly used harmonization dataset iHarmony4~\cite{dovenet}, the composite images are synthetic, so there are no real $\bm{I}_o$, but we can approximately acquire $\bm{I}_o$ (see Section~\ref{sec:groundtruth_albeo}). 
In the real composite image datasets like RealHM \cite{ssh}, we naturally have the original foreground images $\bm{I}_o$. 
With $\bm{I}_o$, we can use neural rendering model \cite{albedo} to extract the ground-truth foreground reflectance map $\bm{A}_{gt}$, because it matches the foreground reflectance in the ground-truth harmonized image. We observe that with the ground-truth foreground reflectance as auxiliary input, the reflectance-guided harmonization network could better predict the ground-truth harmonization result. \par

In practice, we may not have $\bm{I}_o$ and thus $\bm{A}_{gt}$ is unavailable. To address this issue, we design a novel diverse reflectance generation network solely based on the foreground. Specifically, the network takes in a foreground and a random vector, producing multiple plausible  foreground reflectances $\bm{A}_{pre}$ by sampling random vectors.  
By using different $\bm{A}_{pre}$ as the auxiliary input of harmonization network, we can obtain multiple plausible harmonization results. 
Extensive experiments demonstrate that our reflectance generation network can generate diverse and reasonable $\bm{A}_{pre}$, and the reflectance-guided harmonization network can produce multiple plausible harmonization results which are consistent with the variation of $\bm{A}_{pre}$. \par

In summary, our major contributions can be summarized as follows,
\begin{itemize}
\item We explore diverse image harmonization considering uncertain foreground reflectance. 
\item We propose a reflectance-guided harmonization network, which reveals that ground-truth foreground reflectance can help approach ground-truth harmonized result. 
\item We design a diverse reflectance generation network to produce diverse and reasonable foreground reflectances, leading to multiple reasonable harmonization results.
\end{itemize}

\section{Related Works}
\subsection{Image Harmonization}
Image harmonization can be broadly categorized into traditional methods and deep learning based methods. 
Traditional image harmonization methods~\cite{t1_1,t1_2,t2_1,t2_2,t2_3} primarily focus on designing color transformations to match the visual appearance between foreground and background. 
Deep learning based approaches \cite{deepharmonization,ssh,ssam,regionaware,guo2021image,bao2022deep,hang2022scs,RenECCV2022,more1,more2,more3,more4,more5,more6,more7} have become mainstream in the field of image harmonization. DoveNet~\cite{dovenet} introduced the first large-scale iHarmony4 dataset. IIH~\cite{IIH}  decomposed composite images into reflectance and illumination components, and performed self-supervised training. iS$^2$AM~\cite{iSSAM} proposed the integration of S$^2$AM~\cite{ssam} module to better capture the relation between the background and foreground. Recently, CDTNet~\cite{CDTNet}, DCCF~\cite{DCCF}, PCT-Net~\cite{PCT-Net}, and other methods~\cite{Harmonizer,WangCVPR2023,LEMaRT,S2CRNet} utilized deep learning networks to predict transformation coefficients, achieving further enhancement in performance while efficiently scaling to high-resolution images. \par 
However, all the above harmonization methods assume that the harmonization result is deterministic, which is problematic. Our method is the first to generate multiple plausible harmonization results given a composite image.

\subsection{Diverse Image-to-Image Translation}
Diverse image-to-image translation aims to transform an input image in the source domain to multiple possible outputs in the target domain. 
We focus on supervised diverse image-to-image translation, which provides one ground-truth output for each input image. For example, BicycleGAN \cite{bicycle} combined cVAE-GAN~\cite{cvae-gan1,cvae-gan2,cvae-gan3} and cLR-GAN~\cite{clr-gan1,clr-gan2,clr-gan3}, allowing the generation of multiple plausible results for a single input image while also mitigating the mode collapse problem. PixelNN~\cite{pixelnn} employed the nearest-neighbor method to combine pixel matching, thereby converting incomplete conditional inputs into multiple outputs. Subsequently, many studies have been done for the diverse image generation tasks. For example, PiiGAN~\cite{piigan} used an additional style extractor. PUT~\cite{PUT} utilized a patch-based vector quantized variational auto-encoder~\cite{vqvae} and an unquantized transformer. PICNet~\cite{PICNET} refined cVAE~\cite{cvae-gan1} to fit the inpainting task, while ICT~\cite{ICT} improved generative performance through Gibbs sampling and transformers.

In our task, we have one ground-truth harmonization result for a composite image, but there could be multiple plausible results, which falls into the scope of supervised diverse image-to-image translation. We are the first to explore diverse image harmonization.

\begin{figure*}[t]
\centering
\includegraphics[width=1.0\linewidth]{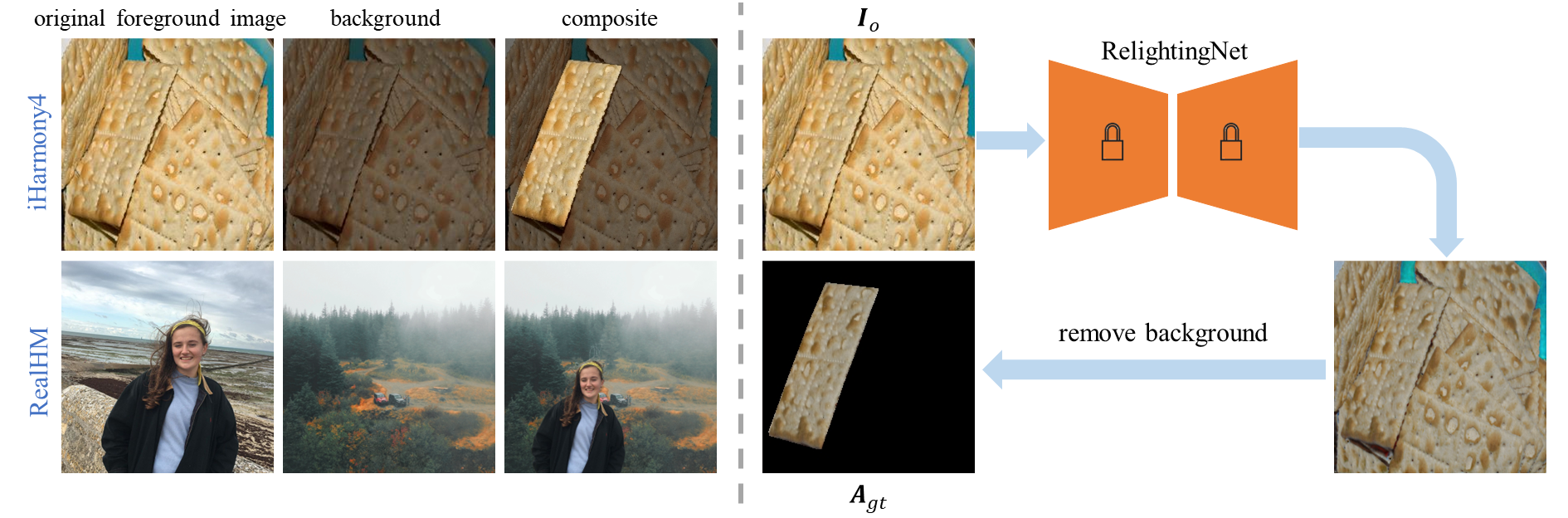}
\caption{In the left part, we show two example triplets of original foreground image, background, and composite image on iHarmony4 \cite{dovenet} and RealHM \cite{ssh}. In the right part, we use pretrained RelightingNet \cite{albedo} to extract the ground-truth foreground reflectance map $\bm{A}_{gt}$ from the original foreground image $\bm{I}_o$.}
\label{fig:getalbedo}
\end{figure*}

\section{Our Method}
Given a composite image $\bm{I}_c$ and foreground mask $\bm{M}$, our goal is generating multiple plausible harmonization results. According to Retinex theory~\cite{kimmel2003variational}, an image $\bm{I}$ can be decomposed into reflectance map $\bm{A}$ and illumination map $\bm{L}$: $\bm{I}= \bm{A} * \bm{L}$,
where $*$ is element-wise product. In a composite image, the foreground reflectance map is independent of the background and estimable based on the foreground region. Solely based on the foreground region, the foreground reflectance map could have multiple possibilities, and different foreground reflectance maps would lead to different harmonization results.

If we have the foreground reflectance map matching the ground-truth harmonized foreground, which is referred to as ground-truth foreground reflectance map $\bm{A}_{gt}$, we expect it could help produce the harmonization result closer to the ground-truth. Therefore, we adapt the existing harmonization network to accept the foreground reflectance map as an auxiliary input, guiding the generation of harmonization results.
We will introduce how to extract ground-truth foreground reflectance map in Section~\ref{sec:groundtruth_albeo} and investigate the effectiveness of using ground-truth foreground reflectance map as guidance in Section~\ref{sec:reflectance_harmonization_network}. 
n. 

In practice, the ground-truth foreground reflectance map $\bm{A}_{gt}$ is often unavailable, so we design a diverse reflectance generation network to produce multiple plausible  foreground reflectance maps $\bm{A}_{pre}$ based on the composite foreground, which will be detailed in Section~\ref{sec:reflectance_generation_network}. 
Finally, the generated foreground reflectance maps can be fed into our reflectance-guided harmonization model to produce multiple plausible harmonization results. 

\subsection{Ground-truth Reflectance Extraction} \label{sec:groundtruth_albeo} 

When we only have the composite foreground, the illumination information is often ambiguous, and thus its foreground reflectance map is also ambiguous. However, if we possess the original foreground image $\bm{I}_o$ that the foreground belongs to, we can  infer the overall illumination more easily based on $\bm{I}_o$ with complex semantics and colors, thereby determining the ground-truth foreground reflectance map $\bm{A}_{gt}$. Next, we will discuss how to get $\bm{I}_o$ from the existing harmonization datasets.

The iHarmony4 \cite{dovenet} dataset is the most commonly used large-scale dataset for image harmonization, comprising four sub-datasets constructed in different ways. iHarmony4 is a synthetic dataset, in which the foregrounds of real images are adjusted to create synthetic composite images. Thus, there are no original foreground images that the composite foregrounds belong to. However, we can simulate the original foreground images $\bm{I}_o$ for experiments. 

The four sub-datasets in iHarmony4 can be divided into two groups. The first group contains HCOCO and HFlickr, in which the composite foregrounds are adjusted from real foregrounds using traditional color transfer methods. We can apply the same color transfer to the entire real image to simulate $\bm{I}_o$.  The second group contains HAdobe5k and Hday2night, in which each scene is associated with a set of images (captured over time or retouched by photographers). The foreground in one image $\bm{I}$ is superseded by that in another image $\bm{I}'$ to create a composite image $\bm{I}_c$, so we can take $\bm{I}'$ as the original foreground image $\bm{I}_o$. 
For all subdatasets, simulated $\bm{I}_o$ are essentially obtained by making global illumination adjustments towards the ground-truth real image, so the foreground reflectance map in $\bm{I}_o$ matches that in the ground-truth image.

Besides the synthetic dataset, we also consider the real dataset RealHM \cite{ssh} consisting of real composite images, in which the composite foregrounds are cropped from one image and pasted on another background image, so we naturally have the original foreground images $\bm{I}_o$ that composite foregrounds belong to. The ground-truth images in RealHM are obtained by manually adjusting the composite foregrounds. We conjecture that when manually adjusting the composite foregrounds, the original foreground images are observed to better estimate the foreground reflectance map. Thus, the foreground reflectance map in $\bm{I}_o$ matches that in ground-truth image.  \par

\begin{figure*}[t]
\centering
\includegraphics[width=1.0\linewidth]{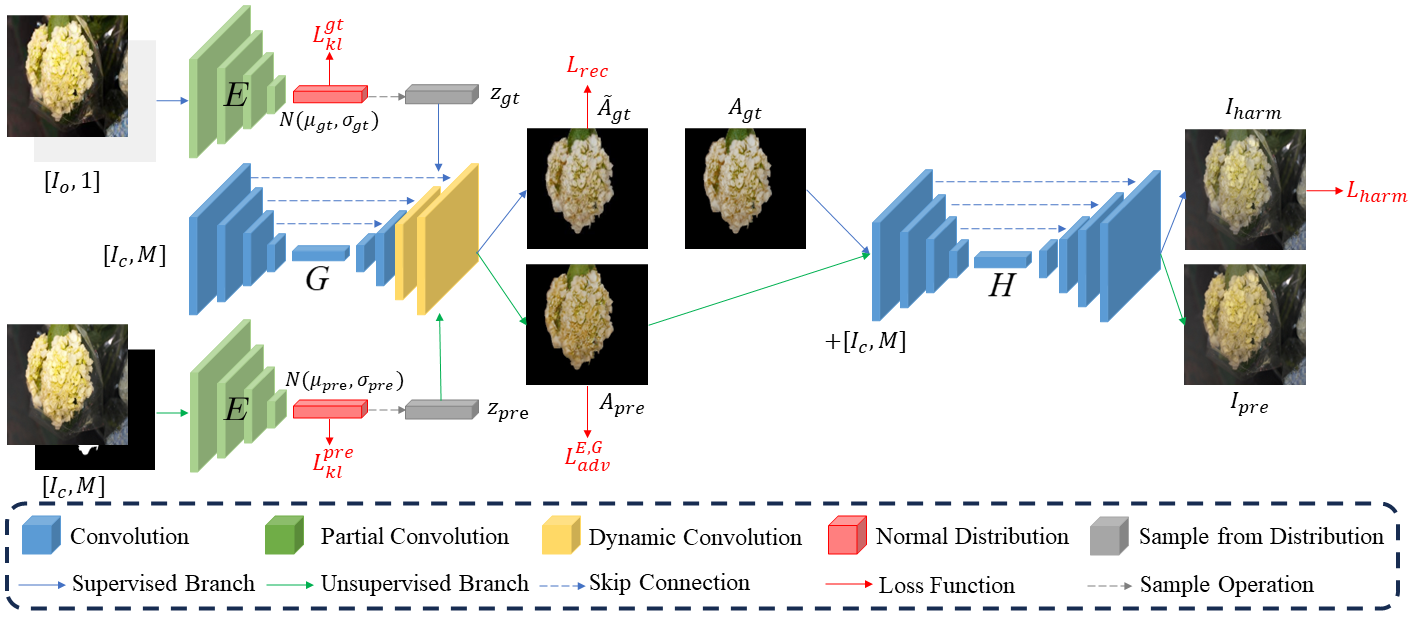}
\caption{The left part shows the diverse reflectance generation network with two branches: supervised and unsupervised branch. In the supervised branch, we extract the guidance code from $\bm{I}_o$, which guides the network $G$ to reconstruct the ground-truth foreground reflectance map. In the unsupervised branch, we extract the guidance code from the foreground region, which guides $G$ to predict plausible foreground reflectance maps. The right part shows the reflectance-guided harmonization network, in which foreground reflectance map is appended to the input to guide the harmonization process. }
\label{fig:network}
\end{figure*}

In summary, for both synthetic dataset iHarmony4 and real dataset RealHM, we can get simulated or real original foreground images $\bm{I}_o$, in which the foreground reflectance map matches that in ground-truth image. Then, we use off-the-shelf reflectance prediction model to extract the reflectance map from  $\bm{I}_o$. 
Specifically, we employ the inverse neural rendering model RelightingNet \cite{albedo}, which can decompose an image into a reflectance map and an illumination map. Although the results are not perfectly accurate, we observe that most predicted reflectance maps are reasonable. 
Figure \ref{fig:getalbedo} provides a detailed illustration of the various forms of $\bm{I}_o$ for different datasets and the process of getting $\bm{A}_{gt}$, which will be used in Section~\ref{sec:reflectance_harmonization_network} and \ref{sec:reflectance_generation_network}.

\subsection{Reflectance-Guided Image Harmonization} 
\label{sec:reflectance_harmonization_network}
We conjecture that provided with ground-truth foreground reflectance map $\bm{A}_{gt}$, the uncertainty of harmonization process could be alleviated, and the harmonization result should be closer to the ground-truth image. In other words, the ground-truth foreground reflectance maps could serve as auxiliary information for arbitrary harmonization network to improve the performance.

Considering that the existing image harmonization methods have quite diverse network structures,  
we opt for the simplest strategy to inject ground-truth foreground reflectance map into network without severely distorting the network structure. 
Specifically, we concatenate $\bm{A}_{gt}$ with the original input $[\bm{I}_c, \bm{M}]$ channel-wisely, which is sent to the harmonization network to produce the harmonized image $\bm{I}_{harm}$. 
Different harmonization methods also have different loss functions. For simplicity, we opt to maintain the original loss function for each method, which is denoted as $L_{harm}$.

\begin{table*}[t]
\small
\centering
\setlength{\tabcolsep}{1mm}{%
\begin{tabular}{c|c|c|c|c|c|c|c|c|c|c|c|c|c|c|c|c}
\hline
 \multirow{2}{*}{Backbone} & \multirow{2}{*}{Method} & \multicolumn{3}{c|}{All}  & \multicolumn{3}{c|}{HCOCO} & \multicolumn{3}{c|}{HFlickr} & \multicolumn{3}{c|}{HAdobe5k} & \multicolumn{3}{c}{Hday2night} \\  
\cline{3-17}
 ~ & ~ & MSE & fMSE & PSNR & MSE & fMSE & PSNR & MSE & fMSE & PSNR & MSE & fMSE & PSNR & MSE & fMSE & PSNR
 \\ \hline
 \multirow{4}{*}{iS$^2$AM} & base & 24.64 & 262.67 & 37.95 & 16.48 & 266.14 & 39.16 & 69.68 & 443.63 & 33.56 & 22.59 & 166.19 & 37.24 & 40.59 & 591.07 & 37.72 \\
 ~ & fg & 25.45 & 277.68 & 38.03 & 17.70 & 285.08 & 38.94 & 70.83 & 457.96 & 33.48 & 21.62 & 163.08 & 38.06 & 54.81 & 778.49 & 36.85 \\
 ~ & gt & \textbf{21.37} & \textbf{231.01} & \textbf{38.59} & \textbf{14.62} & \textbf{240.61} & 39.49 & \textbf{54.90} & \textbf{344.73} & \textbf{34.42} & \textbf{20.75} & \textbf{148.60} & \textbf{38.47} & 39.87 & \textbf{552.47} & 37.79 \\ 
 ~ & pred(10) & 22.94 & 243.51 & 38.58 & 15.79 & 251.28 & \textbf{39.51} & 61.84 & 381.17 & 34.27 & 21.29 & 155.99 & 38.46 & \textbf{38.05} & 557.82 & \textbf{37.80} \\ \hline
 \multirow{4}{*}{PCT-Net} & base & 18.16 & 216.25 & 39.85 & 10.72 & 208.26 & 40.78 & 44.30 & 341.10 & 35.13 & 21.25 & 157.24 & 39.97 & 44.74 & 654.81 & 37.65 \\
 ~ & fg & 22.11 & 261.06 & 38.95 & 13.25 & 257.87 & 39.88 & 56.62 & 401.92 & 34.36 & 24.87 & 184.64 & 38.97 & 48.15 & 728.09 & 37.48 \\
 ~ & gt & \textbf{16.93} & \textbf{199.41} & 39.92 & \textbf{10.62} & \textbf{198.14} & 40.76 & \textbf{40.46} & \textbf{279.18} & \textbf{35.76} & 19.06 & \textbf{144.63} & \textbf{40.00} & \textbf{38.98} & 633.27 & 37.68 \\
 ~ & pred(10) & 17.72 & 209.45 & \textbf{39.96} & 11.51 & 208.21 & \textbf{40.86} & 43.09 & 303.87 & 35.71 & \textbf{18.67} & 150.01 & 39.94 & 44.18 & \textbf{627.06} & \textbf{37.87} \\
 \hline
\end{tabular}
}
\caption{Quantitative comparison on iHarmony4 dataset. The best results for each backbone network are denoted in boldface.}
\label{tab:results_iHarmony4}
\end{table*}

\subsection{Diverse Reflectance Generation} 
\label{sec:reflectance_generation_network}
However, in practical applications, we usually do not have access to the original foreground image $\bm{I}_o$ and thus cannot get $\bm{A}_{gt}$. In this case, we only have the composite foreground and its foreground reflectance map is uncertain. Therefore, we aim to generate multiple reasonable foreground reflectance maps for the composite foreground. To reach this goal, we design a novel diverse reflectance generation network with two branches: supervised branch and unsupervised branches. 
Both branches share the reflectance generation U-Net, which takes in the composite foreground and a random vector to produce a foreground reflectance map. The random vector serves as the guidance code. We can get multiple foreground reflectance maps by sampling the guidance code multiple times. However, the distributions of guidance codes in two branches are different.
In the supervised branch, the guidance code follows the encoded distribution from $\bm{I}_o$. Under the guidance of $\bm{I}_o$, the predicted foreground reflectance map is expected to approach $\bm{A}_{gt}$. In the unsupervised branch, the guidance code follows the encoded distribution from $\bm{I}_c$, making it usable at test time when $\bm{I}_o$ is unavailable.
Next, we will introduce reflectance generation U-Net and the two branch separately.

\subsubsection{Reflectance Generation U-Net} \label{sec:reflectance_UNet}

We adopt U-Net $G$ in \cite{iSSAM}, which takes in composite foreground $[\bm{I}_c, \bm{M}]$ and predicts the foreground reflectance map. Various approaches have been explored in previous works to inject a guidance code $\bm{z}$ into the network. 
For more effective utilization of $\bm{z}$, following \cite{styleganv2}, we use $\bm{z}$ to predict dynamic kernels, which act upon the decoder feature maps in the U-Net. 
Specifically, we first pass $\bm{z}$ through several fully connected layers to get $\bm{w}$, which contains the input channel weights of the dynamic convolution kernel. The convolution kernel modulated by $\bm{w}$ is applied to the final feature map in the decoder, and the resultant feature map accounts for predicting the foreground reflectance map. For the details of dynamic kernel, please refer to \cite{styleganv2}.

\subsubsection{Supervised Branch} \label{sec:supervised_branch}
In the supervised branch, we have the  original foreground image $\bm{I}_o$. We aim to  reconstruct the ground-truth foreground reflectance map $\bm{A}_{gt}$ under the guidance of $\bm{I}_o$. 
We employ the encoder $E$ to encode $\bm{I}_o$ into the Gaussian distribution $\mathcal{N}(\bm{\mu}_{gt},\bm{\sigma}_{gt})$. 
Subsequently, we sample from this distribution to get the guidance code $\bm{z}_{gt}$ by using reparameterization trick \cite{cvae-gan1}. To regulate the encoded distribution, we introduce a KL divergence loss to enforce $\mathcal{N}(\bm{\mu}_{gt},\bm{\sigma}_{gt})$ to be close to unit Gaussian distribution:
\begin{eqnarray}
L_{kl}^{gt}=KL[\mathcal{N}(\bm{\mu}_{gt},\bm{\sigma}_{gt})||\mathcal{N}(\bm{0},\bm{1})]. 
\end{eqnarray}

$\bm{z}_{gt}$ is injected into the reflectance generation U-Net to predict the foreground reflectance map $\tilde{\bm{A}}_{gt}$, which is pushed towards $\bm{A}_{gt}$. As we are only interested in the foreground region, we adopt a foreground MSE loss $L_{rec}$:
\begin{eqnarray}
L_{rec}=\|\tilde{\bm{A}}_{gt}*\bm{M}-\bm{A}_{gt}\|^2,
\end{eqnarray}
where $*$ means element-wise multiplication.

\subsubsection{Unsupervised Branch} \label{sec:unsupervised_branch}
In the unsupervised branch, we cannot access the original foreground image $\bm{I}_o$. Thus, we use the composite image $[\bm{I}_c,\bm{M}]$ to get the guidance code. 
First, we extend the encoder $E$ by using partial convolutions \cite{partialconv} to extract information from partial image based on the provided mask $\bm{M}$. For the supervised branch, we can use all-one mask, in which case partial convolution reduces to vanilla convolution. Similar to the supervised branch, We employ $E$ to encode $[\bm{I}_c,\bm{M}]$ into the Gaussian distribution $\mathcal{N}(\bm{\mu}_{pre},\bm{\sigma}_{pre})$, and sample the guidance code $\bm{z}_{pre}$ from this distribution. 
Since
$\mathcal{N}(\bm{\mu}_{pre},\bm{\sigma}_{pre})$ should have certain overlap with $\mathcal{N}(\bm{\mu}_{gt},\bm{\sigma}_{gt})$, we add a KL divergence loss to prevent that $\mathcal{N}(\bm{\mu}_{pre},\bm{\sigma}_{pre})$ deviates too far from $\mathcal{N}(\bm{\mu}_{gt},\bm{\sigma}_{gt})$: 
\begin{eqnarray}
L_{kl}^{pre}=KL[\mathcal{N}(\bm{\mu}_{pre},\bm{\sigma}_{pre})||\mathcal{N}(\bm{\mu}_{gt},\bm{\sigma}_{gt})].
\end{eqnarray}

\begin{figure*}[t]
\centering
\includegraphics[width=1.0\linewidth]{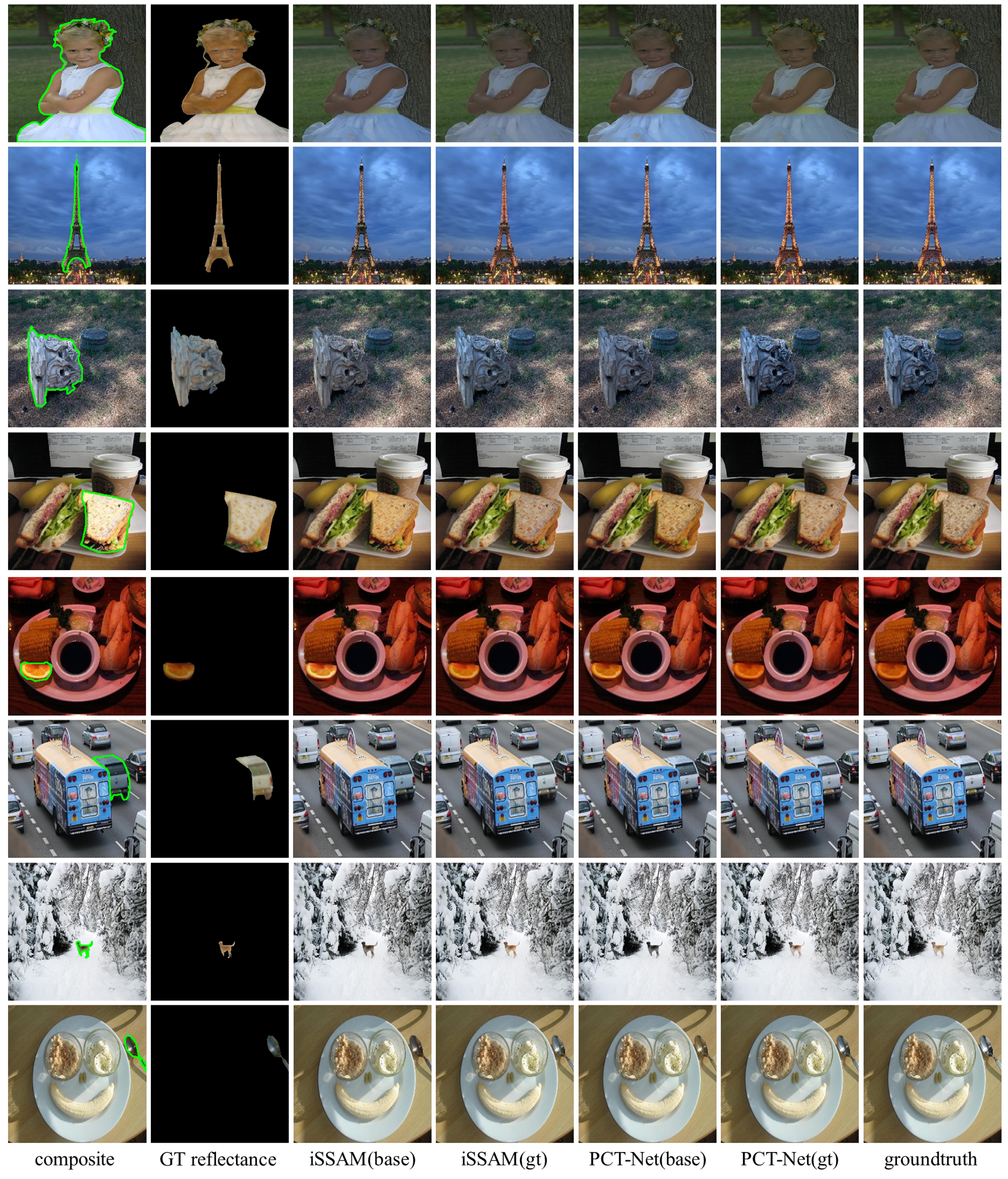}
\caption{The harmonization results of iSSAM~\cite{iSSAM} and PCT-Net~\cite{PCT-Net} on iHarmony4 dataset when using or without using ground-truth foreground reflectance map.}
\label{fig:withreflectance}
\end{figure*}

Then, similar to the supervised branch, we sample the guidance code $\bm{z}_{pre}$ from  $\mathcal{N}(\bm{\mu}_{pre},\bm{\sigma}_{pre})$ and inject it into $G$ to produce the  foreground reflectance map $\bm{A}_{pre}$. 
To ensure the effectiveness of the unsupervised branch, we adopt adversarial learning to make the predicted foreground reflectance maps indistinguishable from real foreground reflectance maps. 
Specifically, we utilize the discriminator in \cite{cgan}, denoted as $D_a$. The discriminator takes in the foreground information $[\bm{A}_{pre},\bm{I}_c*\bm{M}, \bm{M}]$ and predicts the realism score, in which $\bm{I}_c*\bm{M}$ functions as the conditional information to help judge the realism of $\bm{A}_{pre}$. Under the adversarial learning framework, we update the generator $\{E,G\}$ and the discriminator $D_a$ alternatingly. 
When updating the generator $\{E,G\}$, we expect the produced $\bm{A}_{pre}$ to confuse the discriminator $D_a$. We adopt the least-square adversarial loss \cite{lsgan}:
\begin{eqnarray}
L_{adv}^{E,G}=(D_a(\bm{A}_{pre},\bm{I}_c*\bm{M},\bm{M})-1)^2.
\end{eqnarray}
When training the discriminator $D_a$, we expect it to distinguish $\bm{A}_{pre}$ from real foreground reflectance maps.  Hence, the loss function can be written as:
\begin{eqnarray}
L_{adv}^{D_a}=(D_a(\bm{A}_{gt},\bm{I}_c*\bm{M},\bm{M})-1)^2\nonumber \\+D_a(\bm{A}_{pre},\bm{I}_c*\bm{M},\bm{M})^2.
\end{eqnarray}
 
Combining two branches, the total loss function for training $\{E,G\}$ can be summarized as \begin{eqnarray}
L_{E,G}= L_{rec}+L_{kl}^{gt}+L_{kl}^{pre}+\lambda L_{adv}^{E,G},
\end{eqnarray}
in which the hyper-parameter $\lambda$ is set as $0.1$ empirically. 

During test time when the original foreground image is unavailable, we can use the unsupervised branch to produce multiple plausible foreground reflectance maps, which are delivered to the reflectance-guided harmonization network in Section \ref{sec:reflectance_harmonization_network} to generate multiple plausible harmonization results.

\begin{figure*}[t]
\centering
\includegraphics[width=0.96\linewidth]{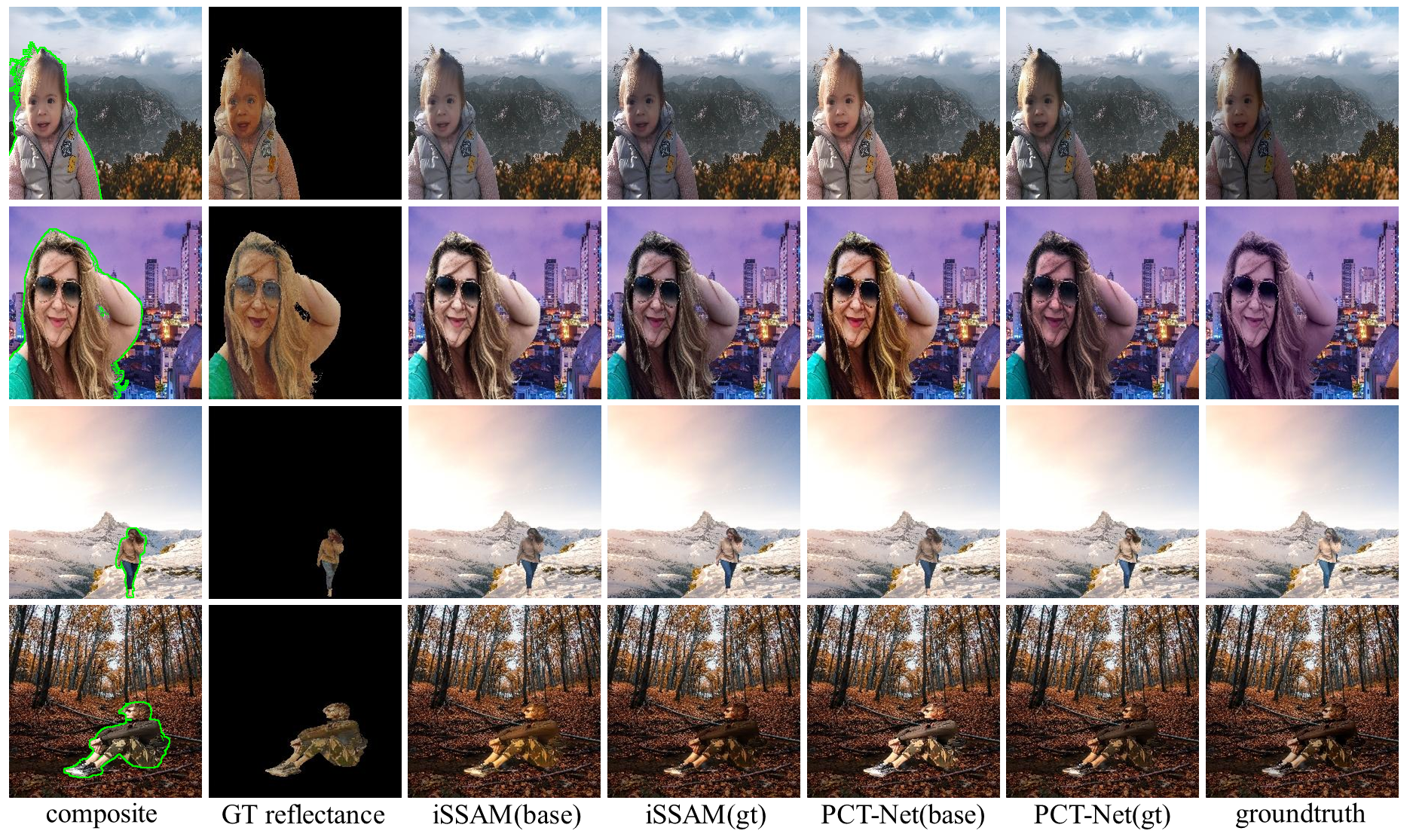}
\caption{The harmonization results of iSSAM~\cite{iSSAM} and PCT-Net~\cite{PCT-Net} on realHM dataset when using or without using ground-truth foreground reflectance map.}
\label{fig:withreflectance_realHM}
\end{figure*}

\section{Experiments}
\subsection{Datasets and Implementation Details}
We conduct experiments on the iHarmony4 dataset with four sub-datasets and the RealHM dataset. Because the RealHM dataset only contains 218 images, we utilize it solely for evaluation. As mentioned in Section~\ref{sec:groundtruth_albeo}, we can get the original foreground images $\bm{I}_o$ and the ground-truth foreground reflectance maps $\bm{A}_{gt}$ for 
these datasets. 

Regarding the harmonization backbone network $H$, we employ iS$^2$AM and PCT-Net as two examples to show the effectiveness of ground-truth foreground reflectance maps in Section~\ref{sec:effectiveness_gt_reflectance}. As the overall performance of PCT-Net is superior to iS$^2$AM, we use PCT-Net by default in the remaining sections unless otherwise stated. Our experimental environment is Ubuntu 18.04, CUDA 11.3, four RTX-3090 GPUs with 24GB memory, and PyTorch 1.10 framework. 
In line with prior works, we adopted Mean Squared Error (MSE), foreground Mean Squared Error (fMSE), and Peak Signal-to-Noise Ratio (PSNR) as evaluation metrics.

\begin{figure*}[t]
\centering
\includegraphics[width=1.0\linewidth]{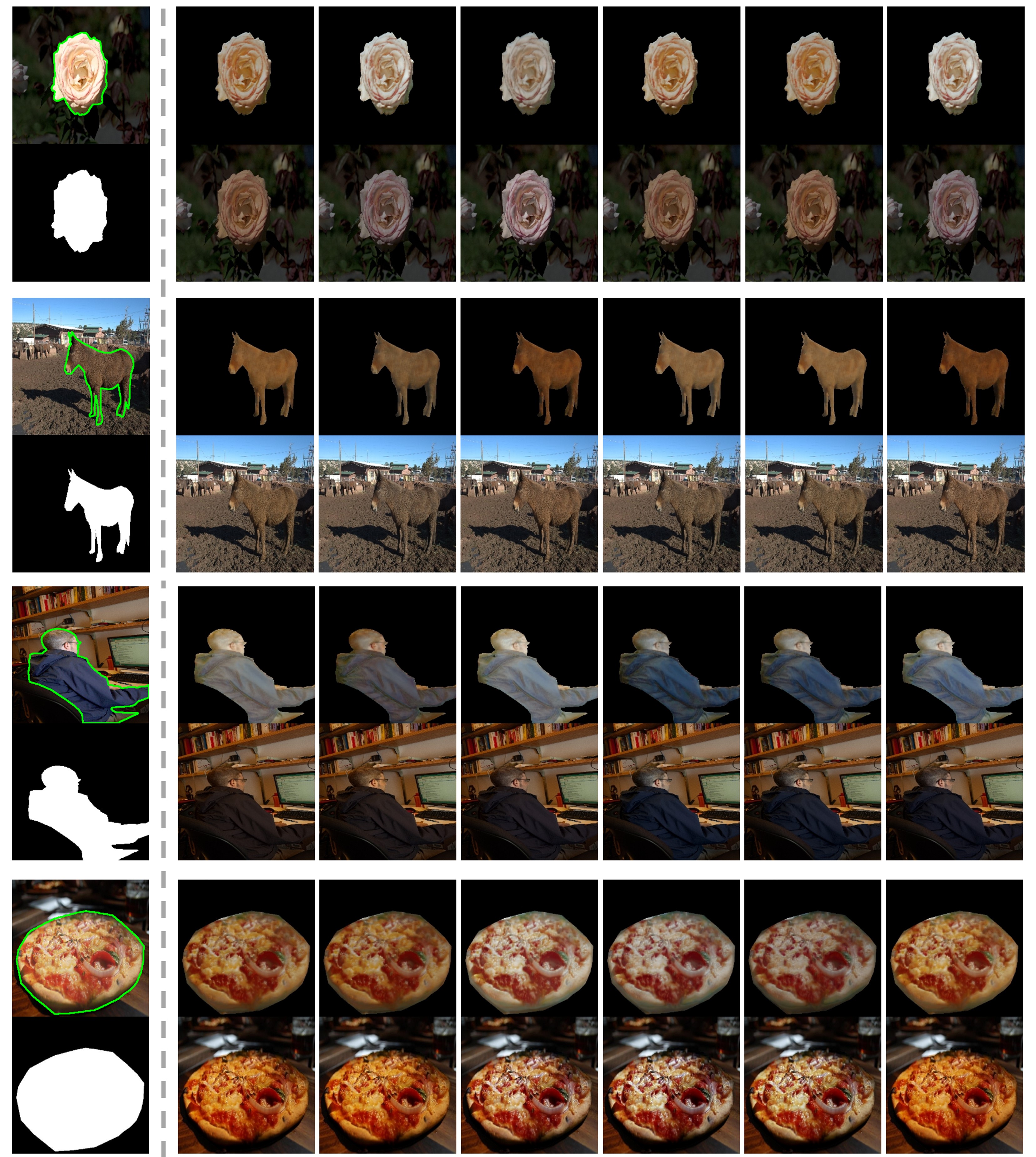}
\caption{Illustration of multiple possible harmonization results arising from the reflectance uncertainty. In each example, the top row represents the predicted foreground reflectance maps, while the bottom row represents the corresponding harmonization result.}
\label{fig:multi}
\end{figure*}

\subsection{Effectiveness of Ground-truth Reflectance}
\label{sec:effectiveness_gt_reflectance}

In this section, we compare the harmonization results of different backbone networks with or without the ground-truth foreground reflectance maps.   

The quantitative outcomes are reported in Table \ref{tab:results_iHarmony4}. In ``base" row, we report the results of the basic backbone network without using ground-truth foreground reflectance maps. In ``gt" row, we report the results of our reflectance-guided harmonization network using ground-truth foreground reflectance maps. 
The comparison between the two rows shows that the integration of ground-truth foreground reflectance maps can make the harmonization results closer to the ground-truth. 

Recall that we use the entire original foreground image to extract the ground-truth foreground reflectance map, because the complex colors and semantics of the entire original foreground image can help alleviate the illumination uncertainty. To validate this point, we extract the foreground reflectance map using only the foreground region and use it in our reflectance-guided harmonization network. The results are reported in ``fg" row. We can see that the performances of ``fg" are much worse than ``gt", and even worse than ``base", indicating that the extracted foreground reflectance map may not match the ground-truth and thus harms the harmonization performance. 

The qualitative results on iHarmony4 are shown in Figure \ref{fig:withreflectance}, from which we can see that including ground-truth foreground reflectance map makes the harmonization results more closely aligned with the ground-truth image.

As analysed in Section~\ref{sec:groundtruth_albeo}, iHarmony4 is a synthetic dataset with simulated original foreground images. In contrast, RealHM contains real original foreground images. 
To further verify the effectiveness of ground-truth foreground reflectance map in real-world cases, we report the results of ``base" and ``gt" on RealHM in Table~\ref{tab:results_RealHM}. 
Since we train the model on iHarmony4 and evaluate on RealHM, the overall results in Table~\ref{tab:results_RealHM} are much worse than those in Table~\ref{tab:results_iHarmony4}, due to the huge domain gap between these two datasets. 
Nevertheless, the results in ``gt" row significantly outperform those in ``base" row, which again validates the effectiveness of ground-truth foreground reflectance map. We also provide qualitative results on RealHM in Figure \ref{fig:withreflectance_realHM}. Although harmonization on RealHM is challenging, the incorporation of $\bm{A}_{gt}$ notably enhances the performance.

\begin{table}[t]
\small
\centering
\setlength{\tabcolsep}{2mm}{%
\begin{tabular}{c|c|c|c|c}
\hline
 Backbone & Method & MSE $\downarrow$& fMSE$\downarrow$ & PSNR$\uparrow$  \\ \hline
 \multirow{2}{*}{iS$^2$AM} & base & 415.61 & 2004.07 & 25.31 \\
 ~ & gt & 294.89 & 1414.42 & 26.46 \\
 \hline
 \multirow{2}{*}{PCT-Net} & base & 385.57 & 1760.69 & 26.08 \\
 ~ & gt & 319.47 & 1493.79 & 26.36 \\
 \hline
\end{tabular}
}
\caption{Quantitative comparison on RealHM dataset.}
\label{tab:results_RealHM}
\end{table}

\subsection{Diverse Harmonization Results} \label{diverse}

In this work, we target at diverse image harmonization. Thus, we show the diversity and plausibility of our approach by providing multiple generated  foreground reflectance maps $\bm{A}_{pre}$ and the corresponding harmonization results in Figure \ref{fig:multi}. Specifically, we first use our diverse reflectance generation network (unsupervised branch) to generate multiple  foreground reflectance maps $\bm{A}_{pre}$. Then, we use $\bm{A}_{pre}$ in our reflectance-guided harmonization network to produce multiple harmonization results. 
In the first example, it is challenging to determine the hue (pure white or a little pinkish) and brightness of foreground illumination. 
Hence, there could be multiple plausible foreground reflectance maps. In the second example, although the horse's color should be brown, its reflectance map could still vary due to the uncertain illumination brightness. 

Furthermore, we can observe a trend of consistency between the harmonization results and the foreground reflectance maps, emphasizing the importance of foreground reflectance in the harmonization process. For instance, a darker reflectance corresponds to a darker harmonization result in both examples. 

We also attempt to analyze the plausibility of multiple harmonization results quantitatively. In particular, for each composite image, we generate $10$ harmonization results and choose the one closest to the ground-truth for metric calculation, leading to row ``pred(10)" in Table \ref{tab:results_iHarmony4}. The results in row ``pred(10)" are better than those in row ``base" and ``fg", and even better than those in row ``gt" in some cases, which proves that our diverse reflectance generation network can produce plausible foreground reflectance maps.

\begin{table}[t]
\centering
\begin{tabular}{c|c|c|c}
\toprule
Row & Method & fMSE$\downarrow$ & LPIPS$\uparrow$ \\
\hline \hline 
1 & w/o supervised branch & 362.81 & 0.0462 \\
2 & w/o unsupervised branch & 297.48 & 0.0832 \\
3 & w/o encoding $[\bm{I}_c,\bm{M}]$ & 264.52 & 0.0587 \\
4 & w/o adversarial training & 259.13 & 0.1165 \\
5 & w/o dynamic conv & 230.17 & 0.0874 \\
6 & full & 200.69 & 0.1031 \\
\bottomrule
\end{tabular}
\caption{Ablation studies on iHarmony4 dataset.}
\label{tab:ablate}
\end{table}

\subsection{Ablation Studies on Our Reflectance Generation Network}
We conduct ablation studies on our diverse reflectance generation network in terms of both plausibility and diversity of the generated foreground reflectance maps. Given a composite image, we sample $10$ times to get $10$ results. For plausibility, we calculate the minimum fMSE compared with the ground-truth one. For diversity, we compute the average of all pairwise LPIPS.

The results are presented in Table \ref{tab:ablate}.
In row 1, we remove the supervised branch and enforce the encoded distribution in the unsupervised branch to approach $\mathcal{N}(\bm{0},\bm{1})$. The generated results are poor and have mode collapse issue. In row 2, we remove the unsupervised branch and sample the guidance code from $\mathcal{N}(\bm{0},\bm{1})$ during testing, leading to a significant performance drop. To further analyze the unsupervised branch, we remove the guidance encoding and adversarial training in row 3 and 4 respectively. Guidance encoding enhances diversity and plausibility, while adversarial training primarily improves plausibility. Lastly, as an alternative way to utilize the guidance code, we directly append the guidance code to the input of $G$ in row 5, which degrades the overall performance.

\section{Conclusion}
In this work, we have studied the uncertainty of harmonization results due to the uncertainty of foreground reflectance. We have designed a reflectance-guided harmonization network and a diverse reflectance generation network. We have demonstrated that the ground-truth foreground reflectance map can benefit the harmonization performance. We have also shown that our method can produce diverse and plausible harmonization results.

{
    \small
    \bibliographystyle{ieeenat_fullname}
    \bibliography{main.bbl}
}

\end{document}